**Transfer learning for cross-modal demand prediction of bike-share and public transit**

Mingzhuang Hua, Ph.D. candidate
School of Transportation, Southeast University
Department of Technology, Management and Economics, Technical University of Denmark
Dongnandaxue Road #2, Nanjing, China, 211189
Email: huamingzhuang@seu.edu.cn

Francisco Camara Pereira, Ph.D., Professor
Department of Technology, Management and Economics, Technical University of Denmark
Bygningstorvet Building 116, Kgs. Lyngby, Denmark, 2800
Email: camara@dtu.dk

Yu Jiang, Ph.D., Associate professor
Department of Technology, Management and Economics, Technical University of Denmark
Bygningstorvet Building 116, Kgs. Lyngby, Denmark, 2800
Email: yujiang@dtu.dk

Xuewu Chen, Ph.D., Professor (Corresponding Author)
Jiangsu Key Laboratory of Urban ITS, Southeast University
Jiangsu Province Collaborative Innovation Center of Modern Urban Traffic Technologies, Southeast University
School of Transportation, Southeast University
Dongnandaxue Road #2, Nanjing, China, 211189
Email: chenxuewu@seu.edu.cn

**Transfer learning for cross-modal demand prediction of bike-share and public transit**

**ABSTRACT**

The urban transportation system is a combination of multiple transport modes, and the interdependencies across those modes exist. This means that the travel demand across different travel modes could be correlated as one mode may receive/create demand from/for another mode, not to mention natural correlations between different demand time series due to general demand flow patterns across the network. It is expectable that cross-modal ripple effects become more prevalent, with Mobility as a Service. Therefore, by propagating demand data across modes, a better demand prediction could be obtained. To this end, this study explores various machine learning models and transfer learning strategies for cross-modal demand prediction. The trip data of bike-share, metro, and taxi are processed as the station-level passenger flows, and then the proposed prediction method is tested in the large-scale case studies of Nanjing and Chicago. The results suggest that prediction models with transfer learning perform better than unimodal prediction models. Furthermore, stacked Long Short-Term Memory model performs particularly well in cross-modal demand prediction. These results verify our combined method's forecasting improvement over existing benchmarks and demonstrate the good transferability for cross-modal demand prediction in multiple cities.

# 1. INTRODUCTION

Compared with only a few decades back, today's transportation system is much more large-scale, heterogeneous, and dynamic. It is heterogeneous because a myriad of new services exists, such as car sharing, ride sharing, shared micro-mobility of bike-share and scooter-share, Mobility as a Service (MaaS) [1]. Autonomous shuttles already operate in several places, and even existing traditional modes, such as metro, bus, or taxi, have their modernized versions, often with smartphone apps, increased electrification, and autonomy. The options for the traveler are certainly more varied today than before. It is also dynamic because these new technologies allow for within-day (sometimes real-time) repositioning/control. Taxi and ride sharing companies often redistribute their fleets during the day, shared micro-mobility and car sharing companies often rebalance at least during the night. Pricing of services can vary by time and zone.

Many cities around the world have operated these new mobility services that can provide users with convenient options, cost-saving benefits, and safe services [2]. As a typical service of micro-mobility, bike-share has been proven to help reduce traffic congestion [3], improve health benefits [4], and protect the environment [5]. These novel mobility services can also solve the first or last mile problems of public transit (metro or bus) [6] and play a key role in urban transportation. In such a dynamic system, the risk of supply-demand misalignment is intuitively greater than in a static system and having accurate demand predictions is vital for efficient responsiveness to demand.

Most existing studies focus on separately predicting real-time demand, e.g., shared micro-mobility and public transit. Thus, even though users can interchange between various transport modes, operators have difficulty in providing collaborative operation of new mobility services and public transit. The key challenge of the collaboration is the lack of demand information exchanging of multiple transport modes [7]. For example, bike-share companies cannot know and then predict metro passenger flow, and metro companies also cannot know and then predict bike-share passenger flow. In the absence of demand information from other transport modes, it is virtually impossible to provide collaborative service and build the MaaS platform. Therefore, it is essential to address the large-scale problem of forecasting demand across different transport modes. This study dedicates to filling this research gap, particularly considering cross-modal predictions between bike-share, metro, and taxi.

The concept of cross-modal predictions demands clarification: it refers to predictions of a certain mode (target) that use information from another mode (source). This includes situations where there is missing data for the target mode but rich data for the source mode (e.g., using metro data to predict bike-share demand soon after it is introduced in a city). It also includes situations of jointly predicting two or more modes given the aggregation of their datasets in a combined model. Therefore, cross-modal prediction is ultimately about data fusion across modes, taking advantage of inter-modal correlations to enhance predictability and data quality.

Of interest is also the concept of *Transfer Learning*, which is the improvement of learning in a new task through the transfer of knowledge from a related task that has already been learned [8]. Transfer learning can help to solve the above two problems because transfer learning methods can input forecasting knowledge into new mobility services such as MaaS from other traditional transport modes, which may have been in operation for many years. There is a research gap in transfer learning among multiple transport modes. In the existing studies of transport demand prediction, transfer learning has gained initial applications, enabling knowledge transfer across time and space. However, these studies [9], [10], [11] focus only on a single transport mode, missing the opportunity of knowledge transfer between different transport modes.

In order to build a transfer learning model among different transport modes, we need to first determine the input-output frames. There are two input-output frames: single-station-input single-station-output (SISO), and multiple-station-input multiple-station-output (MIMO). The SISO frame is popular in demand prediction, and existing MIMO papers are very limited. However, the SISO framework has two insurmountable drawbacks. The first drawback is that SISO consumes too much computing time in large-scale cases. The actual operation should consider hundreds or thousands of stations (or other spatial objects) and predicting the station demand one by one would cost too much time. The second drawback is that SISO is not flexible enough to accommodate cases of different sizes. For example, Wang et al. [9] apply cross-city transfer learning in crowd flow prediction, and the method is match-first-predict-second in the SISO frame. They firstly matched the most similar grid among different cities and then transferred parameters from the similar grid to the target grid. But the SISO method requires that the input amount in different cases be the same, i.e., three cities in Wang et al.'s study all select 400 grids as spatial objects. So, the cross-modal prediction problem should use the MIMO frame. In this study, we establish transfer learning methods that can adapt to different input amounts and adopt the MIMO frame.

There is a trend of collaborative operation for multimodal mobility services in urban transportation, so cross-modal demand prediction is very important in this trend. Besides, demand prediction studies face difficulty in forecasting new services and often suffer from the problem of missing data [12], [13], [14]. Therefore, this paper proposes a cross-modal forecasting framework that incorporates machine learning and transfer learning based on trip data of shared micro-mobility (bike-share) and public transit (metro and taxi). The main contributions of this paper are threefold: (a) Use the cross-modal method to get better prediction results; (b) Establish a transfer learning framework that can handle the data missing problem; (c) Address the difficulty of knowledge transferring in the MIMO models. This study, to the best of our knowledge, is the first paper in cross-modal transport prediction with the transfer learning approach.

The rest of this study is organized as follows. Section 2 introduces related works of machine learning prediction and transfer learning method. Section 3 elaborates the machine learning models and transfer learning strategies in the cross-modal prediction framework. Section 4 describes the data sources and forecasting results of the proposed method. Finally, Section 5 discusses and concludes this study.

## 2. RELATED WORKS

This section reviews existing studies on transport prediction and focuses on two categories, namely, machine learning prediction and transfer learning prediction, as they can be applied to solve the challenging problems in transport forecasting: data missing problem and the inability problem to cope with new mobility services.

### 2.1. Machine Learning for Transport Demand Forecasting

In general, there are two types of machine learning models for transport demand forecasting: single-mode demand model and cross-modal demand model. Single-mode demand model is for a selected transport mode, and cross-modal demand model is for multiple transport modes. As explained before, joint demand model (predicting multiple transport modes in a combined model) is also considered as a part of cross-modal demand modal.

There are many papers discussing single-mode demand models, especially for bike-share and public transit. In bike-share demand prediction, the novel machine learning methods [15], [16] are widely used. Cui et al. [17] use an advanced XGBoost method for flow forecasting and then transform the passenger flow results into bike amount recommendations near a specific metro station. Deep neural networks are also widely used in transport demand forecasting. Zhang et al. [18] count potential bike-share users from public transport and then input this count value into a long short-term memory (LSTM) model to predict bike-share demand. Chai et al. [19] propose a multi-graph convolutional network (GCN) model to predict the station-level bike flow. There are many other studies such as decision trees and neural networks for public transit prediction [20], [21], [22]. Zhang et al. [23] firstly use a GCN and three-dimensional convolutional neural network (CNN) to integrate flow information and forecast metro passenger flow. Zhang et al. [24] take the flow forecast of a single metro station as a multi-input single-output regression prediction problem and use the light gradient boosting machine model to solve this problem. Jia et al. [25] propose an attention-based deep spatio-temporal network with multi-task learning and uses independent channels to model the recent, daily, and weekly metro flows. However, single-mode demand model is based on sufficient data for a transport mode and therefore cannot integrate the data of other transport modes and perform data imputation both for better predictions and for compensating for insufficient/missing data.

Modern urban transport is a large-scale system consisting of multiple transport modes, including new mobility services and traditional public transit [26]. The collaborative service of urban transportation is based on demand prediction in multiple transport modes [27]. Bike-share is a popular type of shared transport, and its correlation with public transit is significant [28]. Meanwhile, bike-share is also a new micro-mobility mode, which starts providing service much later than public transit in many cities [29]. Bike-share data can provide broader spatial information for public transit prediction, and, in turn, public transit data can provide different demand features for bike share prediction. Hence, the cross-modal demand prediction of bike-share and public transit could better understand their relationships and address the drawbacks of lacking data. It is necessary to develop a cross-modal demand prediction methodology, which integrates multiple transport modes into a general prediction framework.

Despite the necessity in practical operation, cross-modal demand prediction, or joint demand prediction in multiple transport modes, has not received sufficient attention in the existing studies. There are only few papers discussing joint demand prediction of multimodal urban transport. Toman et al. [30] use a vector autoregression model to estimate the usage proportions of different transport modes and apply a dynamic linear model to predict the total demand of all modes. Their two-stage method could forecast the city-level usage demand of metro, bike-share, taxi, and ride sharing in New York City. But their study takes the whole city as a single spatial object, which is not station-level and cannot be directly used in the actual operation such as bike-share rebalancing and ride sharing matching. Liang et al. [31] also take New York City as the case study and propose a multi-relational spatio-temporal graph neural network for joint demand prediction in metro and ride sharing. Because of the data limit, their forecast horizon or prediction time interval is four hours, which is not short-term and cannot be directly used in the dynamic operation such as ride sharing matching. Their spatial objects are 136 metro stations and 63 ride sharing zones, which is better than taking the whole city as one place but still a small-scale prediction problem. These studies only discuss the small-scale problem of joint demand prediction, which is not practical in large-scale transport operations. Urban transport, especially MaaS, should consider the large-scale network with hundreds of stations or even thousands of stations. Therefore, a large-scale method for joint demand prediction is required, which is also the focus of our paper.

Transport demand forecasting faces the challenge of large-scale networks [32], [33], and cross-modal demand forecasting makes this problem even more complicated. Both shared micro-mobility and public transit have hundreds or even thousands of stations. The demands for these stations need to be predicted and output at the same time. As a prerequisite for cross-modal demand forecasting, the input-output framework of MIMO [34] needs to be given full attention. But most studies focus on SISO prediction, and the research related to MIMO forecasting is relatively insufficient. Therefore, the novel machine learning models that adapt to the MIMO framework and cross-modal prediction need to be established.

## 2.2. Transfer Learning in Related Studies

Transfer learning applies the knowledge of one domain to another related domain, which can provide better performance. Transfer learning has been widely applied in image processing, classification, and prediction. Ravanbakhsh et al. [35] apply Generative Adversarial Nets and a cross-channel approach in abnormal event detection, which use a discriminator as a supervisor in the video processing. Fawaz et al. [36] applied transfer learning in time series classification, and the input data are the sequences with different lengths. Zhang et al. [37] propose an advanced convolutional LSTM network to predict cellular data, in which a clustering algorithm is used to divide a city into different groups, and then a successive inter-cluster transfer learning strategy is applied for enhancing knowledge reuse. But transfer learning does not get enough attention in transport demand predicting. The existing transfer learning for transport research is limited [9] and primarily focuses on road traffic prediction.

Transfer learning has different strategies, including feature extraction [35], fine-tuning [9], [36], [38], and split-brain [39]. Firstly, feature extraction is the basic transfer learning strategy, which keeps the model unchanged and only changes the input and output data. Li et al. [11] applied various transfer learning strategies into machine learning prediction for real-time road traffic and used the three areas of UK Highways England road networks as the case studies. They selected three points as the source and another point as the target in each area, then transferred the forecasting knowledge from source to target. They found that the feature extraction strategy performs best in their single-mode prediction study. Secondly, fine-tuning is more popular as it can overcome the differences between the source and target tasks. Wang et al. [10] combined transfer learning and deep learning to predict road traffic flow and found that the fine-tuning strategy is suitable in their study. Their method transferred the knowledge from data-rich cases into predicting short-term traffic flow in data-strapped cases, which could solve the data lacking problems. Lastly, as for split-brain, Zhang et al. [39] propose the split-brain autoencoders to obtain two disjoint sub-networks and predict one subset from another, which is applied in predicting color and grayscale of image synthesis tasks. Because of the differences in the knowledge transfer framework, the transfer learning strategies suitable for various prediction problems are also different.

Transfer learning is a promising method in cross-modal transport prediction and solving data missing or lacking problems. As for data missing, many studies have demonstrated the reliability and effectiveness of transfer learning in addressing insufficient data [40], [41], [42], [43]. Besides, transfer learning can be used to transfer forecasting knowledge not only between different areas on the same transport mode but also between different transportation modes in the same area. If the forecasting knowledge is transferred from a small-scale mode to a large-scale mode, transfer learning could also save the running time of large-scale demand prediction. Therefore, transfer learning could be a tool for predicting newly-operated micro-mobility services, especially bike-share and e-scooter sharing, using available public transit data. This is explored by this study.

## 3. METHODS

In this study, for the cross-modal demand prediction, machine learning and transfer learning are combined for the forecasting knowledge transfer between bike-share and public transit. We applied a framework of machine-learning-first transfer-learning-second for this cross-modal prediction problem. The process of transferring bike-share knowledge to public transit prediction is as follows. Firstly, we use machine learning to predict the dynamic demand of public transit; Secondly, we apply different transfer learning strategies to transfer the forecasting knowledge of public transit; Thirdly, we build the machine learning models with transferred knowledge to predict the large-scale dynamic demand of bike-share. Using a similar process, the knowledge of bike-share is also transferred to the machine learning prediction of public transit. In what follows, Section 3.1 presents the machine learning part, Section 3.2 describes the transfer learning part, and, finally, Section 3.3 introduces prediction model benchmarks and performance index.

### 3.1. Stacked LSTM Model for Demand Forecasting

In this study, the stacked LSTM method is selected as the machine learning prediction model. LSTM is an elegant type among many RNN models, which has been widely used in transport prediction [44], [45], [46], [47]. The RNN method has the feature of network delay recursion, which could grasp the patterns of dynamic systems. LSTM improves the basic RNN model with internal mechanisms called gates and the memory cell. The LSTM gates consist of forget gate, input gate, and output gate. This LSTM model can solve the problem of vanishing or exploding gradients and better deal with short-term memory conditions.

Neural network depth is generally attributed to the success of many challenging predictions. In particular, stacked Long Short-Term Memory (Stacked LSTM) is defined as an LSTM model comprised of multiple LSTM layers. Stacked LSTM can increase the depth of LSTM neural networks and has a better performance of prediction tasks. Multiple LSTM layers make the model deeper, more accurately earning the description as a deep learning technique. Besides, multiple LSTM layers of a stacked LSTM model could capture the high-dimensional non-linear patterns of transport demand. Meanwhile, various machine learning methods have also been used to compare with the stacked LSTM model and the corresponding transfer learning strategies, whose details are introduced in Section 3.3.

In a stacked LSTM model, there are three types of layers: input layer, hidden layer, and output layer. In the input layer, the spatio-temporal flow data (the actual demand amounts in the former intervals of all stations) are used as the input matrix. As for the hidden layer, several LSTM layers are stacked and fused into the prediction model, and each LSTM layer has many units. In the output layer, the predicted demand amounts in the future interval of all stations are the output results. The architecture of our stacked LSTM model is shown in Figure 1.

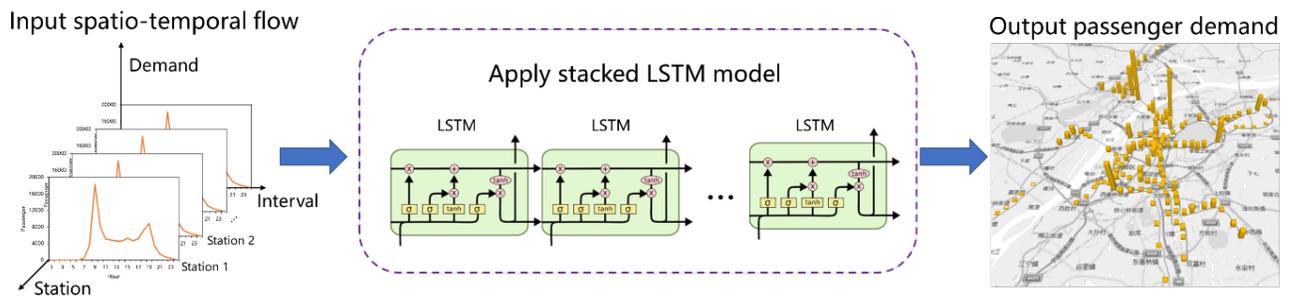

**Figure 1** The prediction approach of stacked LSTM model

## 3.2. Transfer Learning Strategies for Joint Prediction

For the cross-modal demand prediction of bike-share and public transit, a general framework has been established among the various cities and scenarios. The detailed transfer learning frame in cross-modal prediction is proposed as follows. Firstly, the forecast models without transfer of bike-share and public transit are established separately. Secondly, transferring public transit knowledge to bike-share prediction and transferring bike-share knowledge to public transit prediction is to build forecast models with transfer. Lastly, compare the results of the forecast models without and with transfer. The architecture of transfer learning for cross-modal demand prediction is shown in Figure 2.

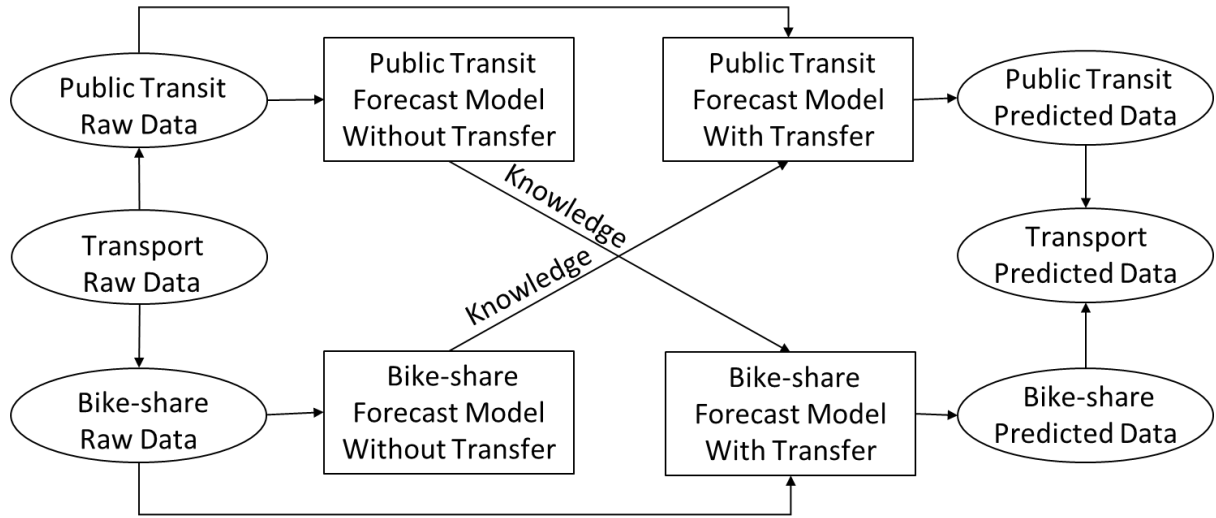

**Figure 2** Transfer learning used in cross-modal demand prediction

There are four transfer strategies: feature extraction (does not change model), fine-tuning without freezing transferred layers (FT), fine-tuning with freezing transferred layers (FTF), split-brain (SB). Transferring parameters is the key to this MIMO prediction problem. In MIMO's Neural Networks, weights of hidden layers could be transferred, but weights of input or output layers cannot be transferred. Besides, the demand magnitudes of bike-share and public transit are different, so there are different input or output amounts in the MIMO frame. Because the input/output of origin and target domains are different, the feature extraction strategy cannot be applied in the cross-modal demand prediction. Therefore, three strategies FT, FTF, and SB are applied in this paper.

Firstly, the details of the FTF strategy are as follows. Because the input and output amounts of bike-share and public transit are different, the input and output layers of the prediction model should be tuned and cannot be frozen. The hidden layers are the transferred layers in transfer learning, which can be tuned or frozen. If the weights of the hidden layers from the related task are frozen in the target task, it is the transfer learning strategy of fine-tuning with freezing transferred layers. Secondly, FT is different from the FTF strategy. If the weights of the hidden layers from the related task are tuned in the target task, it is the transfer learning strategy of fine-tuning without freezing transferred layers. Thirdly, the SB strategy is a novel transfer learning type with two sub-tasks. The split-brain strategy splits the prediction task into two disjoint sub-tasks and predicts the output of one subset with the model of another subset. In the split-brain prediction study of Zhang et al. [39], one sub-task predicts depth from images, while the other predicts images from depth.

**3.3. Baselines and Metrics**

*3.3.1. Baseline Methods*

Five benchmarks are used to compare the forecasting performance of the proposed method, and the brief introduction of these baselines are as follows:

a) One step. It takes the observed value at the former interval as the predicted value at the next interval.

b) Historical average (HA). The average observed value during the past weeks at the same period in time and the same station are calculated as the predicted value.

c) Vector autoregression (VAR). It uses a vector to generalize the autoregressive model and therefore is suitable for MIMO prediction. Its advantage is capturing the interdependencies of multiple time series.

d) Random forest (RF). It consists of multiple decision trees and outputs the average prediction of these trees. A decision tree is a tree-like model that learns the decision rules from root to leaf. The root is the entire sample, the branch is the feature conjunction, and the leaf is the target value.

e) Graph Convolutional Network (GCN) is a type of convolutional neural network that can work directly on graphs, which means that filter parameters are typically shared over all locations in the graph. GNN and CNN are combined to build the GCN model. GCN is good at learning graph representations and has achieved superior performance in many tasks.

*3.3.2. Evaluation Metrics*

Mean Absolute Error (MAE) and root mean square error (RMSE) are taken as the evaluation metrics of model performance. The calculation of MAE and RMSE are displayed in Equations (1), (2), and (3).

$$\text{MAE} = \frac{1}{N \times T} \sum_{i=1}^{N} \sum_{t=1}^{T} |y_{i,t} - \hat{y}_{i,t}| \qquad (1)$$

$$\text{RMSE} = \sqrt{\frac{1}{N \times T} \sum_{i=1}^{N} \sum_{t=1}^{T} (y_{i,t} - \hat{y}_{i,t})^2} \quad (2)$$

$$R^2 = 1 - \frac{\sum_{i=1}^{N} \sum_{t=1}^{T} (y_{i,t} - \hat{y}_{i,t})^2}{\sum_{i=1}^{N} \sum_{t=1}^{T} (y_{i,t} - \bar{y})^2} \quad (3)$$

where $N$ represents the number of stations, $T$ represents the amount of prediction intervals, $y_{i,t}$ represents the observed flow in the $i$-th station in the $t$-th interval, $\bar{y}$ represents the mean value of $y_{i,t}$, and $\hat{y}_{i,t}$ represents the predicted flow in the $i$-th station in the $t$-th interval.

## 4. RESULTS
### 4.1. Dataset Description

The trip data used in this study are from Chicago City and Nanjing City. As shown in Table 1, these two cities both have large-scale multimodal transport services with more than 1,000 spatial objects. The case studies in these two cities are a) bike-share and taxi in Chicago, US; b) bike-share and metro in Nanjing, China. The datasets are used to ensure the methods and findings are not exclusive to a specific scenario. For the Chicago case, trip data in March 2019 of taxi and bike-share are obtained, from the open data website. For the Nanjing case, trip data in March 2019 of metro and bike-share are also obtained from Nanjing Metro Group and Nanjing Public Bicycle Company separately. In March 2019, taxi trips are about ten times bike-share trips in Chicago, and metro trips are about fourteen times bike-share trips in Nanjing. The fields of all four trip datasets contain departure time and position, end time and position.

**Table 1 Trip data of Chicago and Nanjing cases**

| City | Chicago | | Nanjing | |
|---|---|---|---|---|
| Transport mode | Taxi | Bike-share | Metro | Bike-share |
| Trip amount | 1,515,941 | 165,611 | 37,918,284 | 2,786,634 |
| Vehicle amount | 4,691 | 4,331 | - | 50,963 |
| Station amount | 511 | 590 | 160 | 1,475 |
| Travel time (min) | 13.9 | 16.8 | 29.7 | 18.4 |

Figure 3 shows the spatial distributions and temporal characteristics of Chicago and Nanjing transport. For taxi analysis, Chicago city is divided into five hundred small zones. Taxi zone centroids, bike-share stations, and metro stations are considered as the spatial objects of this study. In Chicago, taxi and bike-share services have a similar spatial distribution. As for Nanjing, some bike-share stations are near to metro, but others are far from the metro stations.

Hourly passenger flows of the typical stations at Chicago and Nanjing are also displayed in Figure 3 (c) and (d). In Chicago, the morning peak is 9 to 10 am, and the evening peak is 5 to 6 pm. The bike peak is synchronized with the taxi peak. In Nanjing, the morning peak is 9 am, and the evening peak is 3 to 7 pm. The bike evening peak is earlier than the metro evening peak. The reason for the peak difference may be that the major users of Nanjing Public Bicycle are elderly people who would go grocery shopping from 3 to 4 pm. But metro users mainly travel for commuting, so the metro peak is 7 pm.

The correlation matrix of the station-level demands for taxi and bike-share in Chicago is shown in Figure 3 (e), and the correlation matrix of the station-level demands for metro and bike-share in Nanjing is shown in Figure 3 (f). In these two subfigures, "t" stands for taxi, "b" for bike share, and "m" for metro. In Chicago, the passenger flows of a taxi centroid and other taxi centroids are positively correlated. The passenger flows of a bike-share station and other bike-share stations are also mostly positively correlated. But the passenger flows of some taxi centroids and some bike-share stations are negatively correlated. It reflects the competitive characteristic of taxi and bike-share. The correlation results in Nanjing are different from those in Chicago. In Nanjing, not only are there mostly positive correlations within the same transport mode, but the passenger flows of metro stations and bike-share stations are also basically positively correlated. It reflects the cooperative characteristic of metro and bike-share and the collaborative potential of MaaS.

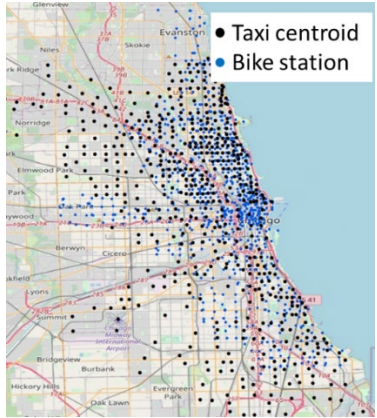
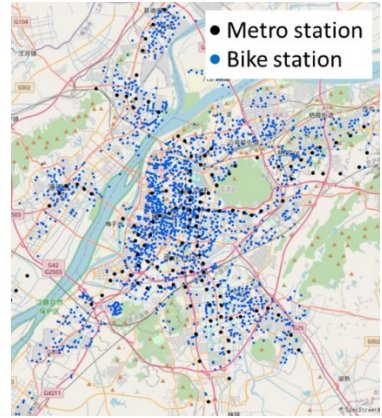

(a) Taxi & bike-share services in Chicago

(b) Nanjing metro & bike-share services

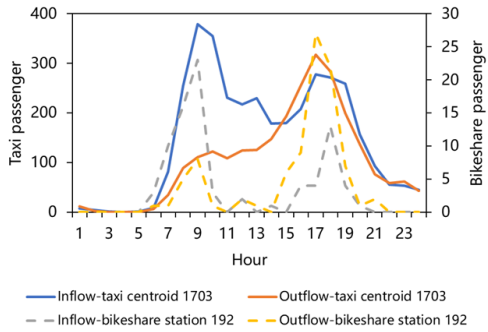
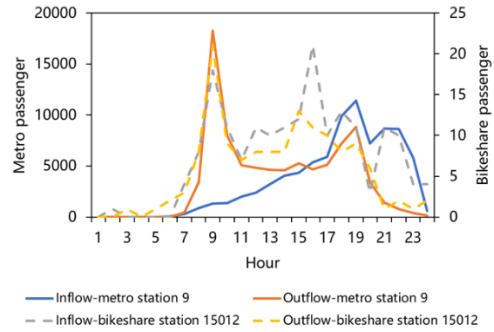

(c) Passenger flow of a taxi centroid and its nearby bike station in Chicago

(d) Passenger flow of a metro station and its nearby bike station in Nanjing

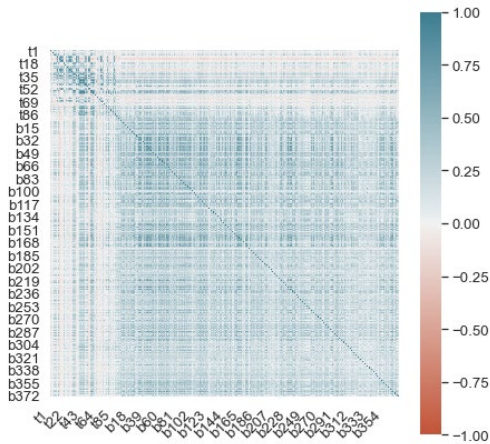
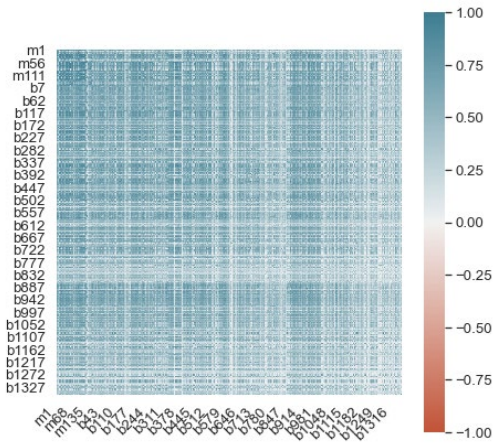

(e) Correlation matrix of station-level demands for taxi and bike-share in Chicago

(f) Correlation matrix of station-level demands for metro and bike-share in Nanjing

**Figure 3** Public transit and bike-share services in Chicago and Nanjing

### 4.2. Experiment Settings

The models of this study are coded with TensorFlow in Python, and their target is to predict the inflow values at all stations in the next horizon. Before estimating model results, the experiment setting needs to be determined, which consists of the spatial object, time interval, training, and test datasets. Firstly, the spatial object in each case would be reduced. The reason for spatial object reduction is that some stations have very low passenger flows. For example, many hourly flow values of these stations are zero. If the average hourly flow is less than 0.1, this station would be removed from spatial objects. After spatial object reduction, the case studies still are large-scale networks. The Chicago case has 374 bike-share stations & 88 taxi centroids, and the Nanjing case has 1361 bike-share stations & 159 metro stations. Secondly, four-time intervals are chosen by trip data conversion. Trip data processing would convert trip data into passenger flow data and is conducted in SQL Server Management Studio with three processing steps. a) Bike-share stations, metro stations, and taxi centroids are selected as spatial objects. b) four forecast horizons or time intervals are set for the dynamic prediction, including 15 min, 30 min, 45 min, 60 min. c) Passenger flows at each station in each horizon are calculated and output. Thirdly, this paper uses the first 70% data as the training dataset, the following 10% data as the validation dataset, and the last 20% as the test dataset in all machine learning prediction models.

Hyperparameters in the proposed models include input horizon amount, hidden layer amount, layer unit amount, training epoch, and batch size. To achieve a balance between underfitting and overfitting, hyperparameter tuning is based on the validation dataset. After hyperparameter tuning, four horizons are selected as the suitable input horizons, as shown in Figure 4 (a). It means that the model input is the inflow values at all stations in the past four horizons. In the stacked LSTM model, the objective loss function is MAE, and the optimization algorithm uses Adam. As shown in Figure 4 (b), the model performs the best when the value of batch size is set to be 150. Generally, the model performance keeps improving as the training epoch increases. The model results become stable when the training epoch reaches 200, which is therefore set as the value of the training epoch. Besides, the hidden layer amount and unit amount of each layer are two important and related hyperparameters that build the basic architecture of the neural network. Grid search is used to optimize these two hyperparameters, and 3 stacked LSTM layers and 100 units in each hidden layer provide the best performance.

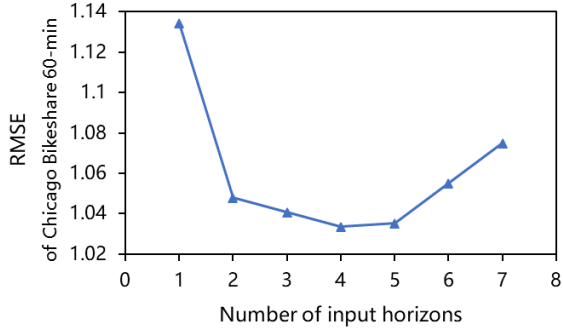 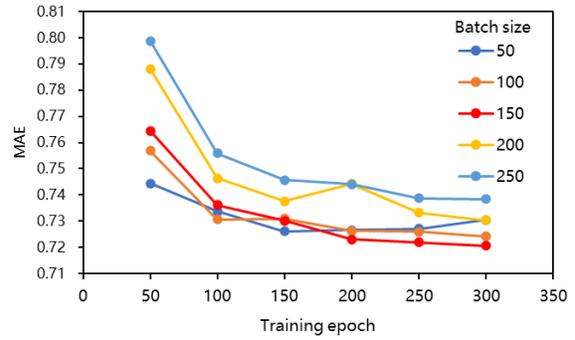

(a) Input horizon amount            (b) Batch size and training epoch

**Figure 4 Hyperparameter selection**

### 4.3. Prediction Results

Table 2 summarizes the prediction results of different transfer learning strategies for Chicago bike-share and taxi. Table 3 summarizes the prediction results of different transfer learning strategies for Nanjing bike-share and metro. For all strategies, the MAE values of public transit are much bigger than bike-share. The reason is that public transit has much more trips than bike-share in both Chicago and Nanjing. It can be found that fine-tuning without freezing transferred layers (FT) strategy has the best performance among all transfer learning strategies. Fine-tuning without freezing transferred layers (FTF) strategy performs slightly worse than FT strategy. This is because changes to the parameters of the transferred layers can further improve the prediction model for different tasks. These findings also suggest that both public transit and bike-share modes can get prediction benefits from the transferred knowledge of each other. Transfer learning is proved to be effective in cross-modal demand prediction of bike-share and public transit.

In order to deal with the missing data problem, there are two solution frameworks discussed in this paper. The first solution is using longer-term data of one transport mode to build a more reliable model and then predicting the demand of another mode. In this study, we also used three-month and six-month passenger flow data to build the corresponding models. The results of these two models did not improve significantly compared to the model based on one-month passenger flow data. Therefore, the solution of longer period data is not applicable. The second solution is to directly use the split-brain (SB) strategy, with one mode of transport passenger flow as input and another mode passenger flow as output. SB strategy is found to be useful in missing-data prediction.

For Chicago and Nanjing bike-share, prediction results of SB strategy are relatively good. It shows that if the data of large-scale spatial objects (bike-share) are missing, small-scale spatial objects (public transit: taxi or metro) can be used as a substitute for predicting the transport mode with large-scale spatial objects.

**Table 2 Transfer learning results in Chicago**

a) Chicago Bike-share

| Period | LSTM-Base | FT | | FTF | | SB | |
|---|---|---|---|---|---|---|---|
| | MAE | MAE | Improving rate | MAE | Improving rate | MAE | Improving rate |
| 15min | 0.206 | 0.161 | **21.8%** | 0.165 | 20.0% | 0.168 | 18.4% |
| 30min | 0.322 | 0.300 | **6.9%** | 0.307 | 4.8% | 0.318 | 1.2% |
| 45min | 0.474 | 0.431 | **9.1%** | 0.437 | 7.9% | 0.459 | 3.2% |
| 60min | 0.613 | 0.555 | **9.4%** | 0.562 | 8.3% | 0.590 | 3.8% |

b) Chicago Taxi

| Period | LSTM-Base | FT | | FTF | | SB | |
|---|---|---|---|---|---|---|---|
| | MAE | MAE | Improving rate | MAE | Improving rate | MAE | Improving rate |
| 15min | 2.497 | 1.623 | **35.0%** | 1.724 | 31.0% | 3.424 | -37.1% |
| 30min | 2.596 | 2.379 | **8.3%** | 2.460 | 5.2% | 4.896 | -88.6% |
| 45min | 4.120 | 3.311 | **19.6%** | 3.454 | 16.2% | 6.788 | -64.8% |
| 60min | 5.835 | 4.606 | **21.1%** | 4.622 | 20.8% | 8.386 | -43.7% |

**Table 3 Transfer learning results in Nanjing**

a) Nanjing Bike-share

| Period | LSTM-Base | FT | | FTF | | SB | |
|---|---|---|---|---|---|---|---|
| | MAE | MAE | Improving rate | MAE | Improving rate | MAE | Improving rate |
| 15min | 0.502 | 0.465 | **7.4%** | 0.473 | 5.7% | 0.473 | 5.8% |
| 30min | 0.774 | 0.724 | **6.5%** | 0.729 | 5.8% | 0.751 | 3.0% |
| 45min | 1.023 | 0.935 | 8.6% | 0.932 | **8.9%** | 1.001 | 2.2% |
| 60min | 1.275 | 1.141 | **10.5%** | 1.149 | 9.9% | 1.284 | -0.7% |

b) Nanjing Metro

| Period | LSTM-Base | FT | | FTF | | SB | |
|---|---|---|---|---|---|---|---|
| | MAE | MAE | Improving rate | MAE | Improving rate | MAE | Improving rate |
| 15min | 13.391 | 12.231 | **8.7%** | 12.628 | 5.7% | 17.625 | -31.6% |
| 30min | 27.998 | 23.190 | **17.2%** | 23.885 | 14.7% | 31.006 | -10.7% |
| 45min | 51.102 | 38.636 | **24.4%** | 40.186 | 21.4% | 43.509 | 14.9% |
| 60min | 74.500 | 57.725 | **22.5%** | 59.047 | 20.7% | 57.780 | 22.4% |

As shown in Tables 4 and 5, the LSTM-FT model shows pretty good performance in the flow prediction for bike-share, and the RF model is suitable for flow prediction of public transit (taxi or metro). Meanwhile, one-step and HA perform worse than the other benchmark models, which indicates the traditional models are not good at predicting dynamic and nonlinear spatio-temporal demands. The VAR model performs very poorly in the dynamic demand prediction of bike-share but shows pretty good performance for the dynamic demand prediction of public transit. This finding reflects that the autoregressive time series model can effectively describe the passenger flow patterns of public transit, but it is not suitable for bike-share services with drastic changes in demand. The performance of the GCN model is not outstanding, which is not consistent with the existing research. The possible reason is that the GCN model needs to input more spatial and temporal data, such as land use and weather conditions, to build a better spatio-temporal GCN model. In short, our proposed method shows good performance in most short-term forecasting scenarios.

**Table 4 Prediction Results of different models in Chicago**

a) Chicago Bike-share

| Model | 15min | | | 30min | | | 45min | | | 60min | | |
|---|---|---|---|---|---|---|---|---|---|---|---|---|
| | MAE | RMSE | $R^2$ | MAE | RMSE | $R^2$ | MAE | RMSE | $R^2$ | MAE | RMSE | $R^2$ |
| One Step | 0.236 | 0.648 | -0.207 | 0.409 | 0.967 | 0.039 | 0.561 | 1.255 | 0.141 | 0.722 | 1.611 | 0.123 |
| HA | 0.197 | 0.540 | 0.160 | 0.337 | 0.804 | 0.335 | 0.454 | 1.019 | 0.433 | 0.558 | 1.228 | 0.491 |
| VAR | 0.290 | 0.550 | 0.128 | 0.536 | 0.930 | 0.111 | 0.824 | 1.402 | -0.072 | 1.236 | 2.086 | -0.469 |
| RF | 0.211 | **0.484** | **0.325** | 0.350 | **0.720** | **0.467** | 0.467 | **0.927** | **0.532** | 0.571 | **1.140** | **0.561** |
| GCN | 0.224 | 0.489 | 0.311 | 0.374 | 0.742 | 0.434 | 0.499 | 0.962 | 0.496 | 0.598 | 1.176 | 0.533 |
| LSTM-FT | **0.161** | 0.525 | 0.206 | **0.300** | 0.788 | 0.362 | **0.431** | 1.016 | 0.438 | **0.555** | 1.217 | 0.500 |

b) Chicago Taxi

| Model | 15min | | | 30min | | | 45min | | | 60min | | |
|---|---|---|---|---|---|---|---|---|---|---|---|---|
| | MAE | RMSE | $R^2$ | MAE | RMSE | $R^2$ | MAE | RMSE | $R^2$ | MAE | RMSE | $R^2$ |
| One Step | **1.259** | 3.666 | 0.977 | 2.208 | 6.814 | 0.965 | 3.347 | 11.151 | 0.947 | 4.871 | 16.874 | 0.922 |
| HA | 2.296 | 7.112 | 0.915 | 3.169 | 10.101 | 0.923 | 4.039 | 13.156 | 0.926 | 4.872 | 16.008 | 0.930 |
| VAR | 1.311 | **3.393** | **0.981** | **2.195** | **5.883** | **0.974** | 3.246 | 9.047 | 0.965 | 4.587 | 13.365 | 0.951 |
| RF | 1.629 | 4.314 | 0.969 | 2.360 | 6.738 | 0.966 | **3.181** | 9.981 | 0.957 | **4.174** | 14.161 | 0.945 |
| GCN | 1.808 | 3.531 | 0.979 | 2.756 | 6.377 | 0.969 | 3.474 | **8.237** | **0.971** | 4.646 | **11.753** | **0.962** |
| LSTM-FT | 1.623 | 4.098 | 0.972 | 2.379 | 6.681 | 0.966 | 3.311 | 9.697 | 0.960 | 4.606 | 14.086 | 0.946 |

**Table 5 Prediction Results of different models in Nanjing**

a) Nanjing Bike-share

| Model | 15min | | | 30min | | | 45min | | | 60min | | |
|---|---|---|---|---|---|---|---|---|---|---|---|---|
| | MAE | RMSE | $R^2$ | MAE | RMSE | $R^2$ | MAE | RMSE | $R^2$ | MAE | RMSE | $R^2$ |
| One Step | 0.631 | 1.324 | 0.291 | 1.029 | 2.137 | 0.440 | 1.423 | 3.003 | 0.459 | 1.862 | 4.071 | 0.410 |
| HA | 0.546 | 1.086 | 0.523 | 0.829 | 1.582 | 0.693 | 1.055 | 1.988 | 0.763 | 1.255 | 2.358 | 0.802 |
| VAR | 1.067 | 1.807 | -0.322 | 4.415 | 11.869 | -16.271 | 1.921 | 3.220 | 0.378 | 1.933 | 3.248 | 0.625 |
| RF | 0.508 | **0.949** | 0.635 | 0.757 | **1.389** | **0.763** | 0.967 | 1.775 | 0.811 | 1.179 | 2.191 | 0.829 |
| GCN | 0.527 | **0.949** | **0.636** | 0.808 | 1.468 | 0.736 | 1.004 | 1.822 | 0.801 | 1.236 | 2.237 | 0.822 |
| LSTM-FT | **0.465** | 0.966 | 0.622 | **0.724** | 1.401 | 0.759 | **0.935** | **1.769** | **0.812** | **1.141** | **2.134** | **0.838** |

b) Chicago Taxi

| Model | 15min | | | 30min | | | 45min | | | 60min | | |
|---|---|---|---|---|---|---|---|---|---|---|---|---|
| | MAE | RMSE | $R^2$ | MAE | RMSE | $R^2$ | MAE | RMSE | $R^2$ | MAE | RMSE | $R^2$ |
| One Step | 15.516 | 40.269 | 0.939 | 40.690 | 116.483 | 0.871 | 76.811 | 217.980 | 0.794 | 120.272 | 335.085 | 0.720 |
| HA | 12.436 | 36.429 | 0.950 | 21.484 | 68.806 | 0.955 | 30.213 | 100.331 | 0.956 | **38.804** | 131.110 | 0.957 |
| VAR | 10.856 | **23.590** | **0.979** | 20.560 | **44.921** | **0.981** | 35.945 | **77.467** | **0.974** | 54.780 | **116.836** | **0.966** |
| RF | **10.836** | 29.907 | 0.966 | **18.809** | 56.760 | 0.969 | **27.904** | 88.076 | 0.966 | 42.049 | 140.617 | 0.951 |
| GCN | 14.095 | 30.905 | 0.964 | 26.563 | 61.356 | 0.964 | 39.543 | 93.338 | 0.962 | 62.260 | 145.476 | 0.947 |
| LSTM-FT | 12.231 | 33.415 | 0.958 | 23.190 | 68.451 | 0.955 | 38.636 | 108.197 | 0.949 | 57.725 | 154.908 | 0.940 |

## 5. DISCUSSION AND CONCLUSIONS

This study focuses on the cross-modal demand prediction for multiple transport modes, which plays a key part in promoting service cooperation, increasing travel transfer, and improving dynamic operation. Transport demand prediction faces the problems of data missing or lacking and cannot effectively adapt the demand information of other transport modes. To deal with these problems, a combined framework of machine learning and transfer learning is proposed. To be more specific, the stacked long short-term memory model with fine-tuning strategy is established for cross-modal demand prediction. Firstly, suitable spatial objects are selected, and trip datasets are processed into passenger flow data. Secondly, machine learning models are used to build the basic prediction model without knowledge transfer. Lastly, various transfer learning strategies are applied and compared to get the transferred information and better forecasting results. For estimating the model performance, real-world case studies are conducted on bike-share and public transit services in Chicago and Nanjing. Generally, this work provides insights on how to combine machine learning and transfer learning for multimodal demand prediction.

The key findings of this study are as follows: a) The joint framework of machine learning and transfer learning is effective for the large-scale problem of cross-modal demand prediction; b) The demand information of bike-share and public transit could provide valuable transferred knowledge for predicting the passenger demand of each other; c) Fine-tuning without freezing transferred layers strategy performs the best among all transfer learning strategies; d) The split-brain strategy is effective in handling the missing-data problem; e) The stacked LSTM model can be combined with a suitable transfer learning strategy for solving the cross-modal demand prediction problem. Besides, the spatio-temporal distribution and correlation of bike-share and public transit are discovered by the visualization analysis. Bike-share and metro are mainly in the cooperation state, but bike-share and taxi have a certain degree of competitive relationship.

This study of cross-modal demand prediction can be applied in several aspects. Firstly, the transfer learning approach of this study enables demand forecasting when the data of a transport mode is missing. For example, the MaaS management platform may face data transmission failures of bike-share, and only the passenger flow data of public transport is available. In this case, the model of this study can be used to predict the future passenger flow of bike-share to guide the city-wide MaaS management. Secondly, cross-modal demand forecasting can guide the dynamic operation of multiple transportation modes. For example, the increase in metro passenger flow demand can result in dispatching more shared bikes into the vicinity of metro stations to meet users' travel and transfer needs. Thirdly, cross-modal demand forecasting can be applied for collaborative transportation services. The cross-modal demand forecast results can be used to infer the in-vehicle crowded information of public transit and the supply-demand balance of bike-share. And the information can recommend users to take more reasonable travel choices to better use multi-modal transportation services.

The proposed method of cross-modal demand prediction can be further improved or extended in the following directions. Firstly, many new mobility services of MaaS do not have fixed stations, which makes the large-scale problem for cross-modal demand prediction even more challenging. For example, ride sharing, and e-scooter sharing have no physical stations and move freely around the city. These new mobility services lack a default spatial object such as the bike-share station in our study, which makes their demand prediction more difficult. Therefore, a new concept of the clustering-based virtual station can be applied as the spatial object for MaaS demand prediction. Secondly, our cross-modal forecasting models need to be compatible with other mobility service datasets. New mobility services such as car sharing, stationless bike sharing, and shared automated vehicles have similarities and differences with bike-share in spatio-temporal demand distribution. Transferring the cross-modal model of this study into these new mobility services needs more research efforts. Lastly, cross-modal forecasting model should consider incorporating multi-source data, such as land use and weather conditions. It could be helpful to assess the impact of inputting more information and build a spatio-temporal demand forecasting model.